%% file: AICAS25.tex
\documentclass[conference]{IEEEtran}
\IEEEoverridecommandlockouts
\usepackage{cite}
\usepackage{amsmath,amssymb,amsfonts}
\usepackage{algorithmic}
\usepackage{algorithm}
\usepackage{graphicx}
\usepackage{textcomp}
\usepackage{xcolor}
\def\BibTeX{{\rm B\kern-.05em{\sc i\kern-.025em b}\kern-.08em
    T\kern-.1667em\lower.7ex\hbox{E}\kern-.125emX}}

\usepackage{xcolor}
\usepackage{caption}
\usepackage{subcaption}
\usepackage{dsfont}
\usepackage{tabularx}
\usepackage{multirow}
\usepackage{hyperref}
\usepackage{bbm}
\usepackage{array, makecell}

\newcommand{\eg}{\textit{e}.\textit{g}.}
\newcommand{\ie}{\textit{i}.\textit{e}.}    

\usepackage[compact]{titlesec}
\titlespacing{\section}{0pt}{*1}{*1}

\begin{document}

\title{End-to-end fully-binarized network design: from Generic Learned Thermometer to Block Pruning \\
\thanks{This work is part of the IPCEI Microelectronics and Connectivity and was supported by the French Public Authorities within the frame of France 2030.}
}

%

\author{\IEEEauthorblockN{Thien Nguyen and William Guicquero}
\IEEEauthorblockA{\textit{Smart Integrated Circuits for Imaging Laboratory, CEA-LETI} \\
F-38000, Grenoble, France. Email: [vanthien.nguyen, william.guicquero]@cea.fr \\
}
}

\makeatletter
\def\ps@IEEEtitlepagestyle{
  \def\@oddfoot{\mycopyrightnotice}
  \def\@evenfoot{}
}
\def\mycopyrightnotice{
  {\footnotesize
  \begin{minipage}{\textwidth}
  \centering
  Copyright~\copyright~2025 IEEE. Personal use of this material is permitted. However, permission to use this material \\ 
  for any other purposes must be obtained from the IEEE by sending an email to pubs-permissions@ieee.org.
  \end{minipage}
  }
}

\maketitle

\begin{abstract}
Existing works on Binary Neural Network (BNN) mainly focus on model's weights and activations while discarding considerations on the input raw data. This article introduces Generic Learned Thermometer (GLT), an encoding technique to improve input data representation for BNN, relying on learning non linear quantization thresholds. This technique consists in multiple data binarizations which can advantageously replace a conventional Analog to Digital Conversion (ADC) that uses natural binary coding. Additionally, we jointly propose a compact topology with light-weight grouped convolutions being trained thanks to block pruning and Knowledge Distillation (KD), aiming at reducing furthermore the model size so as its computational complexity. We show that GLT brings versatility to the BNN by intrinsically performing global tone mapping, enabling significant accuracy gains in practice (demonstrated by simulations on the STL-10 and VWW datasets). Moreover, when combining GLT with our proposed block-pruning technique, we successfully achieve lightweight (under 1Mb), fully-binarized models with limited accuracy degradation while being suitable for in-sensor always-on inference use cases.
\end{abstract}


\section{Introduction}
Designing light-weight, fully-binarized models \cite{BNN, MeliusNet} is a promising avenue to boost the efficiency of deep neural network (DNN) execution towards always-on resourced-constrained inference \cite{7799716, 9741809, 10067588, atimv2, 10454391}. For instance, the multiply-accumulate (MAC) operation between 1-bit signed weights and activations can be implemented using XNORs and popcounts only, thus enabling advantageous hardware simplification compared to its full-precision counterpart. Besides, it goes without saying that model binarization also offers a 32$\times$ reduction of memory needs for storing model's weights and local activations. However, despite remarkable progress in BNNs, most of the works only focus on the weights and activations while keeping regular input data to cap accuracy loss. Unfortunately, it remains an issue for a concrete hardware implementation on devices that only support 1-bit computations, without the capability of dealing with integer inputs. 
To design small, fully-binarized networks with limited performance loss, we propose the following two contributions:
\begin{itemize}
    \item a Generic Learned Thermometer (GLT) which is an input data binarization being jointly trained with a BNN, enabling DNNs execution with 1-bit computations only; 
    \item a method to gradually replace complex processing blocks in BNNs by lighter grouped convolutions while limiting accuracy drop thanks to a distributional loss.
\end{itemize}
\section{Related works}
\begin{figure}[h]
    \centering
    \includegraphics[trim=0 150 0 0,clip, scale = 0.26]{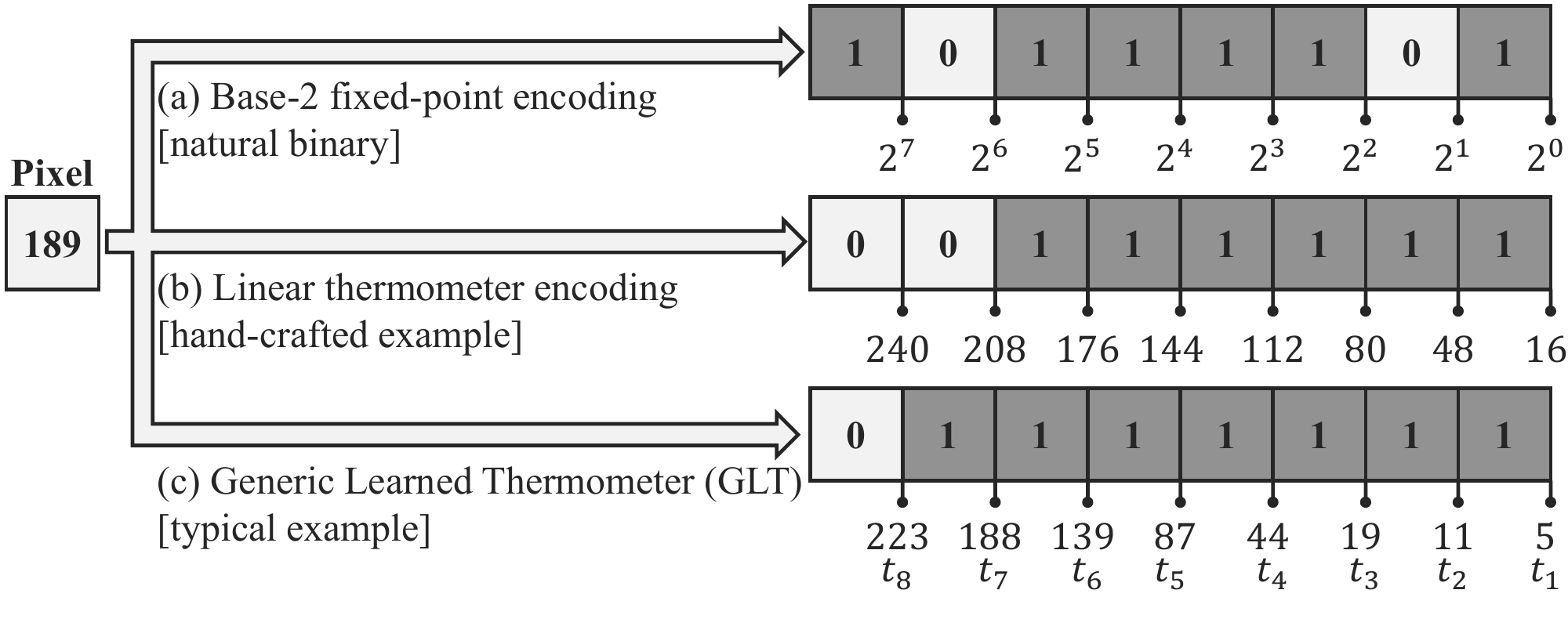}
    \caption{Three examples of encoding techniques for input image data binary representation. Here, for the sake of simplicity, input pixel dynamic range is considered between 0 and 255.}
    \label{stepsize_curves}
     \vspace{-2mm}
\end{figure}

\textbf{Input data binarization:} The de-facto choice to represents input data with a normalized $[0, 1]$ dynamic range is to consider $M$-bit fixed-point coding (\eg, $M$=$8$). In terms of hardware, it thus requires large bit-width multipliers to perform the scalar products, at least at the first layer of the model. To address this issue, \cite{BIL} and \cite{Vorabbi2023InputLB} propose to directly use the base-$2$ fixed-point representation of each input value (Fig.~\ref{stepsize_curves} (a)), with additional depth-wise and point-wise convolutions to combine the resulting bit planes. However, in addition to not being robust to spurious inputs and bit-flips, it is clear that a base-$2$ fixed-point encoding can only be used in the digital domain. Another class of input data binarization is thermometer encoding \cite{8480105}, \cite{FracBNN}. For instance, \cite{FracBNN} introduces a fixed thermometer (FT) encoding with quantization thresholds following a linear ramp (Fig.~\ref{stepsize_curves} (b)).
It is noteworthy mentioning that the thresholds of these thermometric encoding are manually designed and therefore may not be optimal for a wide variety of targeted tasks. On the contrary, GLT allows to automatically learn the thresholds as model's parameters that can be taken into account at the ADC level (Fig.~\ref{nlrADC}) to involve a nonlinear input response, as a global tone mapping.

\textbf{BNNs pruning:} Network pruning \cite{NIPS2015_ae0eb3ee, MLSYS2020_6c44dc73, 9136977, Depthpruning} aims at removing unnecessary operations, operands and weights to reduce model size and complexity. While most existing techniques are mainly designed for full-precision models, only some works address the pruning problem in the context of BNNs. For example, \cite{9136977} employs weight flipping frequency while \cite{10044856} leverages the magnitude of latent weights as the pruning criteria. However, these methods result in irregular structures that require additional hardware features to handle sparse computations. In contrast, our block-based pruning scheme replaces complex blocks by lightweight grouped convolution layers, hence can be directly mapped to canonical hardware platforms. Compared to the existing depth pruning method \cite{Depthpruning}, our method is specifically tailored for BNNs, by gradually pruning complex blocks combining with a distributional loss \cite{liu2020reactnet} to avoid severe accuracy degradation.
\section{Generic Learned Thermometer}
\begin{figure}[h]
    \centering
    \includegraphics[trim=0 150 150 0,clip, scale = 0.3]{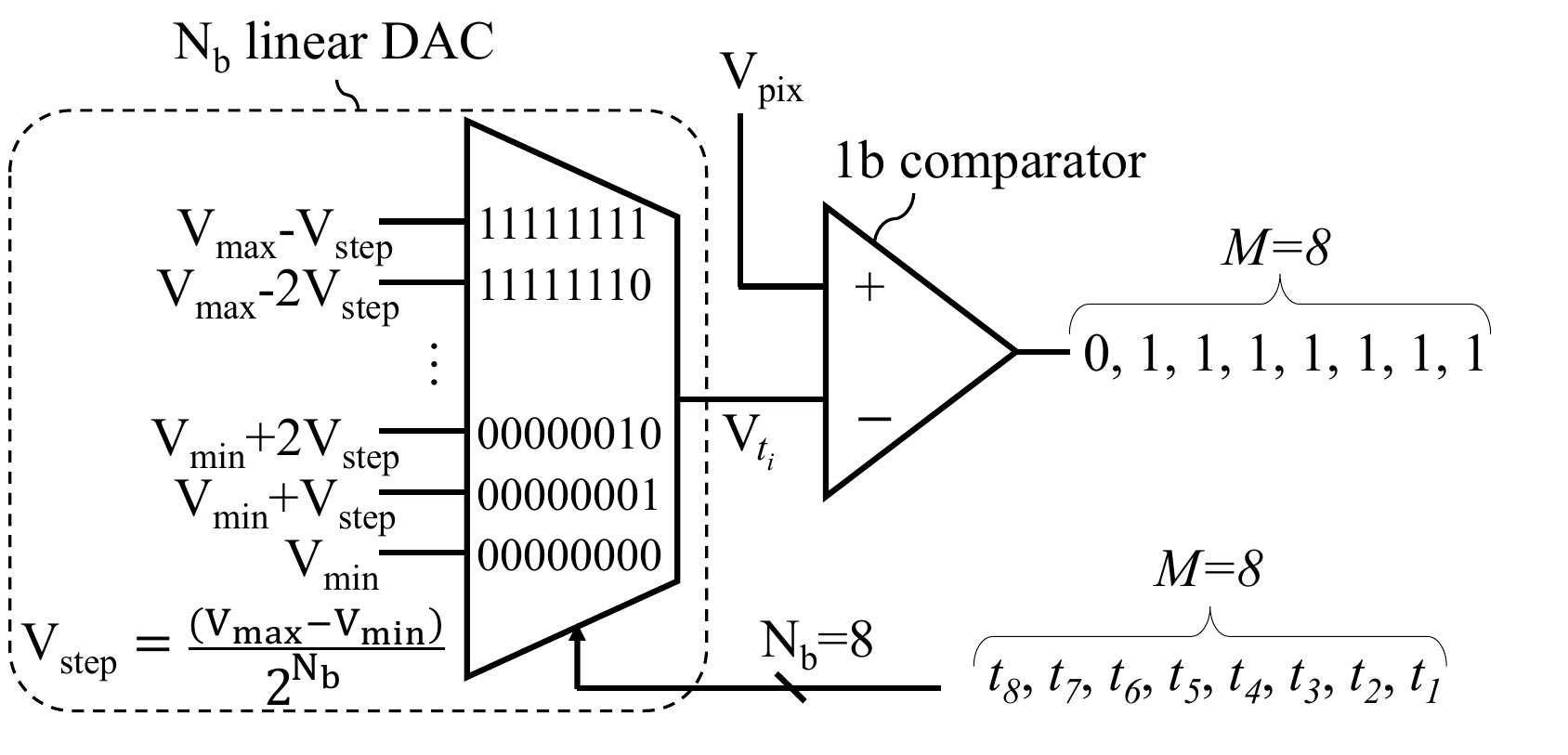}
    \caption{Nonlinear ramp ADC top-level view schematic.}
    \label{nlrADC}
     \vspace{-4mm}
\end{figure}

Our encoding method (GLT) is illustrated in Fig.~\ref{stepsize_curves} (c), which reports a typical example of thermometric thresholds provided as model's parameters, being optimized for a specific inference task via model training. In practice, it intrinsically performs a global tone mapping operation that can be implemented directly during the analog-to-digital conversion stage, if relying on a programmable slope ADC as schematically depicted in Fig.~\ref{nlrADC}. This top-level schematic relies on a 1bit comparator that sequentially outputs comparisons between $\mathrm{V_{pix}}$ (the pixel voltage to convert) and $\mathrm{V}_{t_i}$. For $i \in [\![1,M]\!]$, $\mathrm{V}_{t_i}$ successively takes the $M$ different threshold values $t_i$ encoded on $\mathrm{N_b}$ bits through the use of a DAC composed of a multiplexer and a set of voltage references ranging from $\mathrm{V_{min}}$ to $\mathrm{V_{max}}$. Note that for this example $\mathrm{N_b}$=8 and $M$=8. 
\subsection{Detailed mathematical formulation}
Let us consider an $M$-bit thermometer encoding with $M$ thresholds $0 < t_1 < t_2 < ...< t_{M-1} < t_M < 1$. The thermo-vector $\mathbf{TV} \in \mathbb{R}^{M}$ corresponding to a value $x$ is defined as:
\begin{equation}
TV_i = \textrm{Heaviside}(x - t_i) = \begin{cases}
1 & \textrm{if } x \geq t_i ,\\
0 & \textrm{otherwise}.
\end{cases} 
\label{thermometer_encoding}
\end{equation}
where $x$ is normalized to the interval $[0, 1]$ for ease of later optimization. The choice of the threshold values $\{t_i\}_{i=[\![1,M]\!]}$ deeply impacts the quantity of relevant information preserved and thus model's performance, therefore our goal is to find optimal thresholds by minimizing any task loss. 
Let us denote $\tilde{\mathbf{t}} \in \mathbb{R}_+^{M+1}$ as the learnable latent threshold parameters that are kept always positive during the optimization process. We first compute the normalized vector $\Bar{\mathbf{t}}$ as follows:
\begin{equation}
\Bar{\mathbf{t}} = \frac{\tilde{\mathbf{t}}}{\sum_{i=1}^{M+1}\tilde{t}_i}
\label{normalization}
\end{equation}
With 0 $ <\Bar{t}_i<$1 and $\sum_{i=1}^{M+1}\Bar{t}_i = 1$, the thermometer thresholds $\{t_i\}_{i=[[1, M]]}$ are then computed as a cumulative sum:
\begin{equation}
t_i = \sum_{j=1}^{i} \Bar{t}_j
\label{threshold_compute}
\end{equation}
It can be easily verified that $0 < t_i < t_{i+1} < 1$ for $i \in [\![1,M]\!]$, and the optimal threshold values are indirectly learned through the optimization of $\tilde{\mathbf{t}}$. Let us denote $\mathcal{L}$ as the task loss; $\mathbf{I} \in \mathbb{R}^{H\times W}$ as the input image data of resolution $H\times W$ that is normalized within $[0, 1]$; and $\mathbf{I}^b_i \in \{0, 1 \}^{H\times W}$ as the encoded bit plane at the bit position $i \in [\![1,M]\!]$. Indeed, the gradient with respect to $\tilde{t}_j$ can be written by the chain rule as:

\vspace{-0.1cm}

\begin{equation}
\begin{split}
\dfrac{\partial \mathcal{L}}{\partial \tilde{t}_j} & = \sum_{x, y, i} \left( \dfrac{\partial \mathcal{L}}{\partial I^b_{x, y, i}} \dfrac{\partial I^b_{x, y, i}}{\partial t_i}  \right) \dfrac{\partial t_i}{\partial \tilde{t}_j} \\
& = \sum_{x, y, i} \left( \dfrac{\partial \mathcal{L}}{\partial I^b_{x, y, i}} \dfrac{\partial \textrm{Heaviside}(I_{x, y} - t_i)}{\partial t_i}  \right) \dfrac{\partial t_i}{\partial \tilde{t}_j}
\end{split}
\label{grad}
\end{equation}
where the term $\dfrac{\partial t_i}{\partial \tilde{t}_j}$ is calculated from Eqs. ~\ref{normalization} and ~\ref{threshold_compute}. The gradient vanishing problem caused by Heaviside function could be overcame using approximation techniques \cite{STE, AdaSTE, Wu_2023_ICCV}. Intuitively, near-threshold values should have higher influence (\ie, larger gradient magnitude) than far-threshold counterparts. Therefore, we employ the ReSTE approximation proposed in \cite{Wu_2023_ICCV} and propose a modified version as follows:
\begin{equation}
\dfrac{\partial \textrm{Heaviside}(u)}{\partial u} = \frac{1}{m} \textrm{min}\left( \frac{1}{p} |u|^{\frac{1-p}{p}}, m \right)
\label{ste_grad}
\end{equation}
where $p>1$ controls the narrowness of the bell-shaped gradient curve and $m$ is the clipping threshold. Concretely, $p=2$ and $m=5$ for all of our experiments. Since each $t_i$ is shared for all pixels within the same bit plane, a small change of the latent threshold parameters may notably tamper the model's behavior. Therefore, to stabilize the model convergence during training, the update magnitude of the latent parameters $\tilde{t}_j$ is scaled by a factor $\beta=2/\sqrt{HWM}$.        
\subsection{Latent parameter's initialization and constraint}
An important question is how to properly initialize the latent threshold parameters $\tilde{\mathbf{t}}$. For the sake of simplicity, the straightforward option is to choose the initial values such that the resulting thermometer thresholds follow a linear ramp like proposed in \cite{FracBNN}. In details, the threshold of the $i$-th bit-plane is defined as $t_i=s(i - 0.5)/(2^{\mathrm{N_b}}-1)$ where $s=2^{\mathrm{N_b}}/M$ is the uniform step size and $1/(2^{\mathrm{N_b}}-1)$ is the normalization factor (\eg, 1/255 with 8b image data). Besides, the initialization of the non-normalized latent parameters $\tilde{\mathbf{t}}$ is scaled as: 
\begin{equation}
\tilde{t}_i = \begin{cases}
0.5sk & \textrm{if } i = 1  ,\\
sk    & \textrm{if } 1 < i \leq M, \\  
(0.5s - 1)k    & \textrm{if }  i = M + 1.
\end{cases} 
\label{thermometer_encoding}
\end{equation}
Here the scale factor $k$ aims to balance the stabilization with the updatability of the GLT. For optimization purposes and with respect to training stability issues, the latent parameters are forced to be positive and large enough, \ie, $\tilde{t}_i > 0.05$. Coherently, the scaling value $k$ is set such that the initial latent parameters remain higher than the aforementioned clipping threshold, \eg, by setting $k=M/1280$.   

\section{Block pruning with distributional loss}
\subsection{Proposed method}
\begin{figure}[h]
    \includegraphics[trim=0 50 170 0,clip, scale = 0.3]{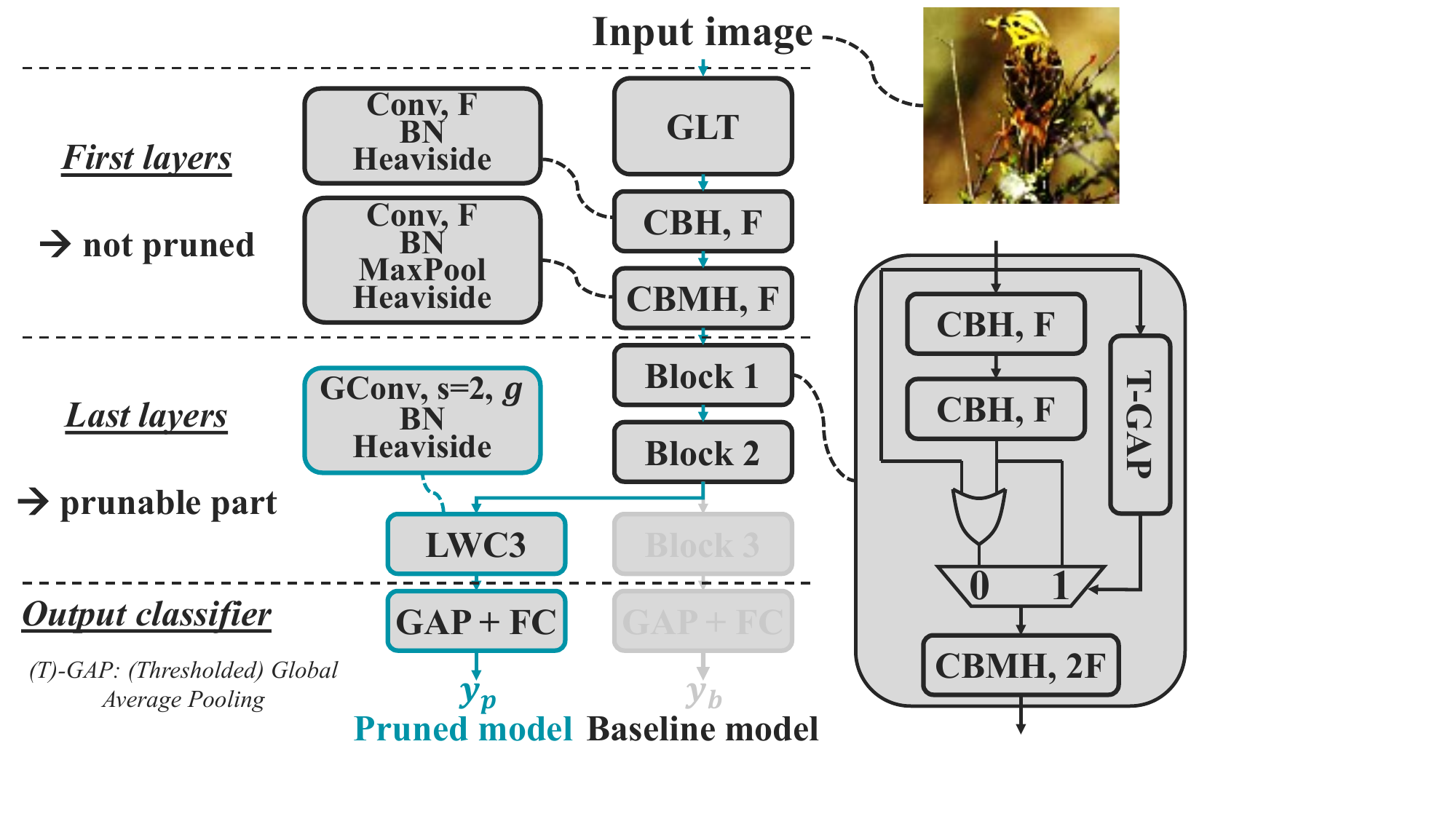} 
    \caption{Block pruning with an auxiliary lightweight grouped convolution (LWC) module. Note that our pruning method can be applied to an arbitrary network.}
    \label{block_pruning}
    \vspace{-3.8mm}
\end{figure}

BNNs usually need to be compressed further to match strict hardware specifications of edge devices, \eg, typically under $1$Mb SRAM. In this section, we propose a method to gradually prune complex processing blocks and replace it with a lightweight convolution (LWC) block involving a $g$-group $3\times3$ convolution (GConv) of strides $2$ followed by a channel shuffle (if $g>1$). Figure~\ref{block_pruning} illustrates how our idea is applied to MUXORNet-11 \cite{MUXORNet}. The number of channels is doubled after the GConv, thus allowing the pruned model to have the same data dimension as the baseline model. The LWC shows a far better hardware mapping than its corresponding block in the baseline model. We leverage the Kullback-Leibler divergence loss forcing the pruned model to learn similar class distributions as the baseline model: 
\begin{equation}
\mathcal{L}_{distr} = \frac{T^2}{N} \sum_{i=1}^N \sum_{j=1}^{\mathcal{C}} y_{b_j} (\mathbf{X}_i; T) \textrm{log} \left(\frac{y_{b_j} (\mathbf{X}_i; T)}{y_{p_j} (\mathbf{X}_i; T)}\right)  
\label{distributional_loss}
\end{equation}
where $y_{b_j} (X_i; T)$ and $y_{p_j} (X_i; T)$ are the softened probability corresponding to class $j$ of the baseline and the pruned model, given the input image $\textbf{X}_i$, \ie, the pre-softmax divided by a given temperature $T$. The total loss function is as follows: 
\begin{equation}
\mathcal{L} = (1 - \lambda) \mathcal{L}_{ce} + \lambda \mathcal{L}_{distr}
\label{total_loss}
\end{equation}
where $\lambda$ is the hyperparameter controlling the balance between the KD loss $\mathcal{L}_{distr}$ and the cross-entropy loss $\mathcal{L}_{ce}$. Concretely, we set $T=8$ and $\lambda = 0.5$ for later experiments. Assume that the model contains $\mathcal{N}_b$ blocks, we gradually prune them in the direction from block $\mathcal{N}_b$ to block $1$. For each stage, we replace each block by a LWC block such that the output dimension is kept unchanged. Without loss of generality, we also notice that here a block may contains a single or several layers. The pruned model is initialized with weights of the previous stage and then retrained with the aforementioned loss function. The complete pruning procedure is presented in Algorithm~\ref{algo}.       

\begin{algorithm}
 \caption{Block pruning algorithm for BNNs.}
 \begin{algorithmic}[1]
 \label{algo}
 \renewcommand{\algorithmicrequire}{\textbf{Input:}}
 \renewcommand{\algorithmicensure}{\textbf{Output:}}
\REQUIRE (1) pre-trained baselines with first layers and classifier models ($\textbf{FL}$ and $\textbf{Cls}$); (2) $\textbf{X}$ the training set; (3) list of $\mathcal{N}_b$ blocks to be pruned $\textbf{W}$ with $\mathcal{P}_b \in [\![1,\mathcal{N}_b]\!]$;
  \STATE $\textbf{W}_p \leftarrow \textbf{W}$
  \FOR {$b = \mathcal{N}_b$ to $\mathcal{P}_b$} 
  \STATE $\textbf{W}_p[b] \leftarrow \textrm{Initialize LWC}_b()$
  \STATE $\textbf{PrunedNet} \leftarrow \textrm{Append}(\textbf{FL}, \textbf{W}_p, \textbf{Cls})$
  \STATE $\textbf{PrunedNet} \leftarrow \textrm{TrainToConvergence}(\textbf{PrunedNet}; \textbf{X})$
  \ENDFOR
  \ENSURE  Optimized pruned model with $\textbf{FL}$, $\textbf{W}_p$ and $\textbf{Cls}$  
 \end{algorithmic} 
 \vspace{-1.5mm}
 \end{algorithm}
 
\section{Experiments}

\subsection{Validation of the GLT} \label{val_glt}
\textbf{Implementation details:} Each channel of the RGB input image is encoded separately into $M$ binary planes (\ie, $M \in \{8, 16, 32\}$), then the resulting binary planes are concatenated together to feed the model. Note that here we only evaluate the effectiveness of the encoding method, therefore we do not consider extra architectures to combine binary planes like proposed in \cite{BIL}, \cite{Vorabbi2023InputLB}. For Visual Wake Words (VWW) dataset \cite{VWW} we resize images to $132\times176$ for training and $120\times160$ for testing. We then perform data augmentation using random cropping to $120\times160$ and random horizontal flipping. For STL-10 dataset \cite{STL10} with $96\times 96$ RGB images, we adopt random crop from all-sided 12-pixel zero-padded images combined with random horizontal flip and random cutout \cite{Devries2017ImprovedRO} of $24\times 24$ patches. We first pre-train the real-valued model with binarized input and ReLU activations, then use it to initialize the weights of the fully-binarized model with 1-b Signed weights and Heaviside activations with STE \cite{STE} gradient. Each model is trained during $100$ epochs for VWW and $250$ epochs for STL-10 with Radam optimizer \cite{Radam}. The learning rate is initialized at $10^{-3}$ and reduced to $10^{-8}$ using cosine decay scheduler.   

\textbf{VWW:} We first conduct a benchmark on the gamma-inversed images ($\gamma=2.2$) of VWW with MUXORNet-11 model. Table~\ref{vww} shows that models with GLT achieve higher accuracy on both train and test sets compared to FT and even the baseline gamma-inversed images in the case of binarized models. Specifically, at $M=16$, we obtain a gap of nearly $0.9\%$ on training and $1.5\%$ on testing. These results demonstrate that GLT allows model to learn more effectively during training and hence generalize better at inference time.          
\begin{table}[h]
\caption{MUXNORNet-11 on gamma-inversed VWW.}
\centering
\renewcommand{\tabcolsep}{4pt}
\begin{tabular}{| c | c | c | c | c | c |}
\hline
\multirow{2}{*}{\makecell{Input encoding \\ method}} & \multirow{2}{*}{\makecell{ $\#$ planes \\ ($M$) }} &\multicolumn{2}{c|}{FP} & \multicolumn{2}{c|}{Bin.} \\
    \cline{3-6}
     &   & Train & Test & Train & Test \\
\hline
Baseline gamma-inversed &  \textit{32-b}  &   96.95  & 91.57     &  89.70   &  87.67 \\ \hline
\multirow{3}{*}{\makecell{Fixed Linear \\ Thermometer \cite{FracBNN}}} & 8 & 96.10    &  89.43    &   90.34    &  86.91 \\ 
  &  16  & 96.37   & 90.51 & 91.25  & 87.39 \\
  &  32  & 96.92   & 90.93 & 92.04  & 88.72 \\
\hline
 \multirow{3}{*}{\makecell{\textbf{Ours} \\ GLT}} & 8 & 96.24   & 90.83  & 90.94  & 88.66 \\
  &  16 &  96.83   & 91.09   &  92.13  &  88.87  \\
   & 32 &  97.17   & 91.23   &  92.39  &  89.43  \\
\hline
\end{tabular}
\label{vww}
\vspace{-3mm}
\end{table}

\begin{figure}
    \centering
    \hspace{-2mm}
    \begin{subfigure}[b]{0.16\textwidth}
        \includegraphics[width=\textwidth]{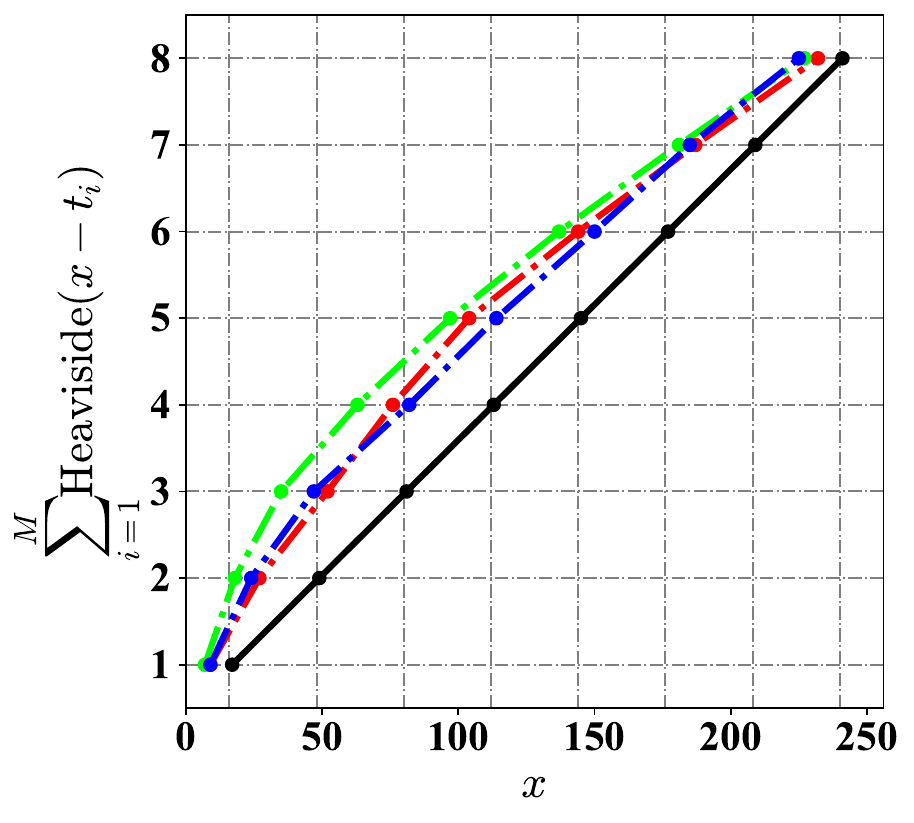}
        \vspace{-6mm}
        \caption{$M=8$}
        \label{vgg_curve}
    \end{subfigure}
    \hspace{-2mm}
    \begin{subfigure}[b]{0.16\textwidth}
        \includegraphics[width=\textwidth]{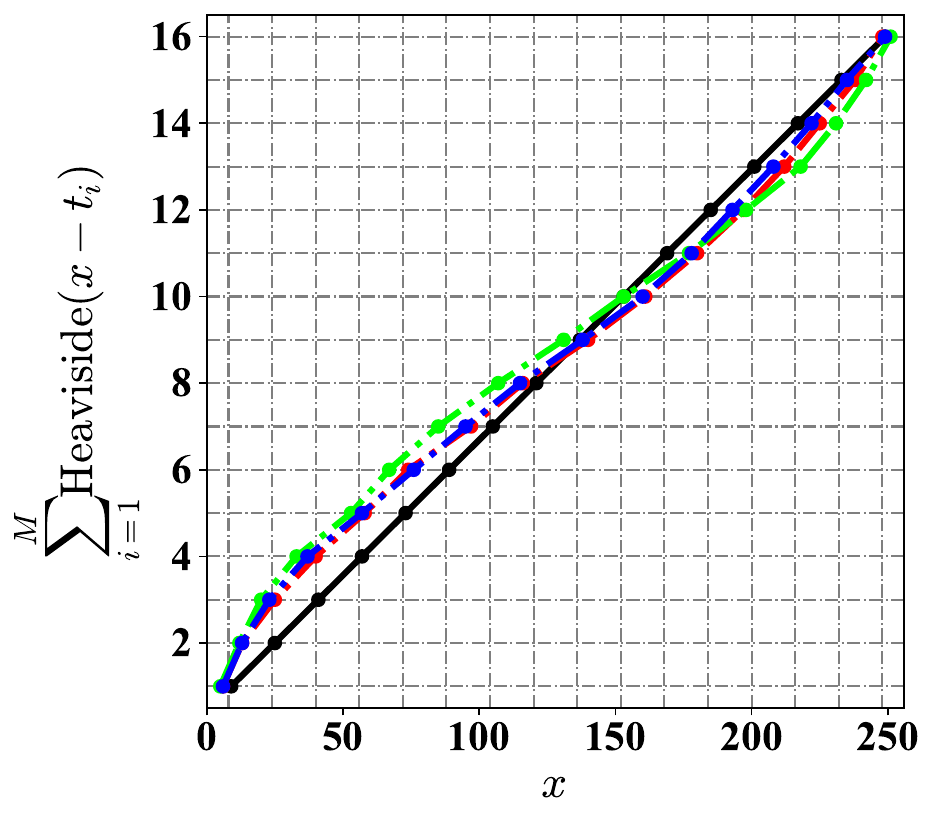}
        \vspace{-6mm}
        \caption{$M=16$}
        \label{vgg_curve}
    \end{subfigure}
    \hspace{-2mm}
     \begin{subfigure}[b]{0.16\textwidth}
        \includegraphics[width=\textwidth]{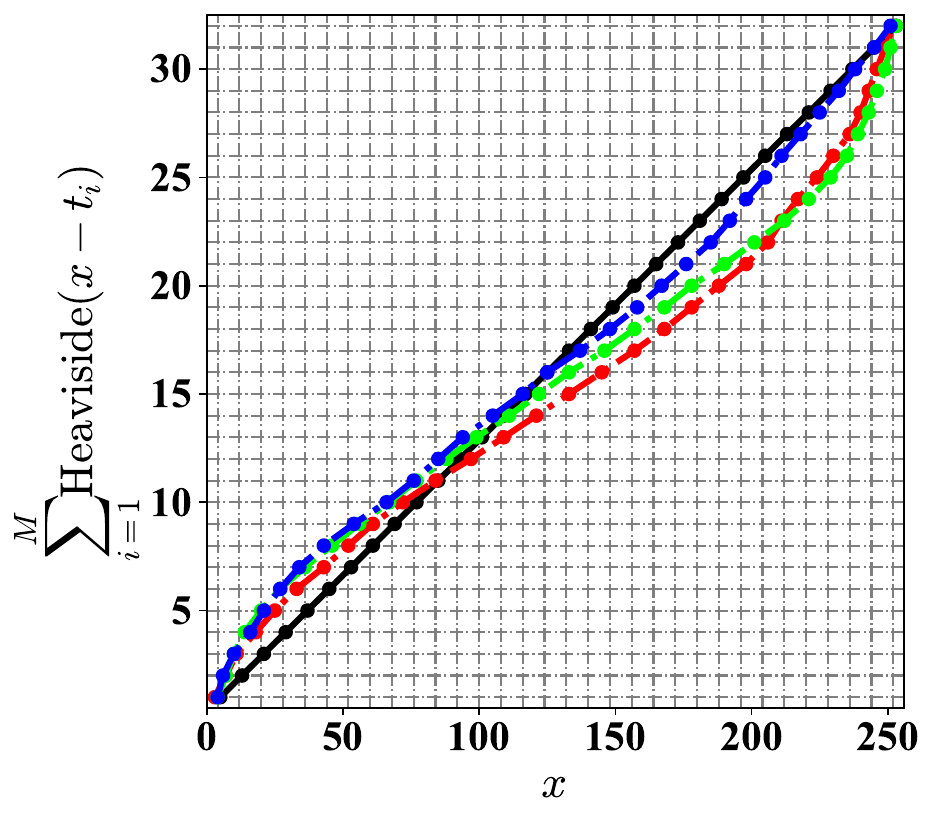}
        \vspace{-6mm}
        \caption{$M=32$}
        \label{muxornet_curve}
    \end{subfigure}
    
    ~
    \vspace{-6mm}
    \caption{Encoding curves of MUXORNet-11 GLT layers trained on gamma-inversed STL-10. \textbf{Black}: fixed linear curve \cite{FracBNN}, \textcolor{red}{\textbf{Red}}/\textcolor{green}{\textbf{Green}}/\textcolor{blue}{\textbf{Blue}}: learned curves of \textcolor{red}{\textbf{R}}/\textcolor{green}{\textbf{G}}/\textcolor{blue}{\textbf{B}} channels.}
    \label{glt_curves}
    \vspace{-6mm}
\end{figure}

\textbf{STL-10:} Table~\ref{encoding_benchmark_digital} and~\ref{encoding_benchmark_gamma_inversed} report the accuracy of models after each stage of training on both the original and the gamma-inversed STL-10 datasets. It is shown that in most cases, model with GLT-encoded input achieves highest accuracy compared to model whose input is encoded by other methods. For the original dataset, at $M=8$, GLT has a slight gain of $1.07\%$ for VGG-Small and $0.85\%$ for MUXORNet-11 compared to FT \cite{FracBNN}. This gain slightly decreases for larger $M$, as more bit planes will increase inter-channel redundancy. On the other hand, for the gamma-inversed dataset, the gain of GLT over FT is strongly boosted up to $2.5\%$ for $M=8$ and it can even retain the accuracy level of FT on the original post-gamma correction dataset. In particular, for $M=32$, MUXORNet-11 with GLT even achieves a slightly higher accuracy than the baseline model with gamma-inversed input. Figure~\ref{glt_curves} shows the curves of the encoding bit-count level (from $1$ to $M$) as a function of the thresholds learned on gamma-inversed dataset. It is shown that although being linearly initialized, our GLT can successfully learn proper nonlinear curves (\ie, global tone mapping), providing a higher accuracy and being well-suited to a real-world deployment scenario. It thus suggests the possibility of  using GLT to perform inference directly from analog data, bypassing all the image rendering stages.   
\begin{table}[h]
\caption{Accuracy ($\%$) on original STL-10 dataset.}
\centering
\renewcommand{\tabcolsep}{4pt}
\begin{tabular}{| c | c | c | c | c | c |}
\hline
\multirow{2}{*}{\makecell{Input encoding \\ method}} & \multirow{2}{*}{\makecell{ $\#$ planes \\ ($M$) }} &\multicolumn{2}{c|}{VGG-Small} & \multicolumn{2}{c|}{MUXORNet-11} \\
    \cline{3-6}
     &   & FP & Bin. &   FP  & Bin. \\
\hline
Baseline integer (8-bit) &  \textit{32-b}  &   83.02   & 79.95     &  84.24   & 79.74 \\ \hline
Base-2 fixed-point \cite{BIL} & 8 & 78.32     & 73.15 &  80.06  & 75.83 \\ \hline
\multirow{3}{*}{\makecell{Fixed Linear \\ Thermometer \cite{FracBNN}}} & 8 & 79.35    &  76.40    &   81.39    &  77.43 \\ 
  &  16  & 80.63   & 77.49 & 82.17  & 78.47 \\
  &  32  & 81.50   & 77.81 & 82.51  & 78.90 \\
\hline
 \multirow{3}{*}{\makecell{\textbf{Ours} \\ GLT}} & 8 & 79.93   & 77.47  & 81.62  & 78.28 \\
  &  16 &  80.87   & 78.02   &  82.73  & 78.60  \\
   & 32 &  81.40   & 78.24   &  82.79 &  79.06  \\
\hline
\end{tabular}
\label{encoding_benchmark_digital}
\vspace{-4mm}
\end{table}

\begin{table}[h]
\caption{Accuracy ($\%$) on gamma-inversed STL-10 dataset.}
\centering
\renewcommand{\tabcolsep}{4pt}
\begin{tabular}{| c | c | c | c | c | c |}
\hline
\multirow{2}{*}{\makecell{Input encoding \\ method}} & \multirow{2}{*}{\makecell{ $\#$ planes \\ ($M$) }} &\multicolumn{2}{c|}{VGG-Small} & \multicolumn{2}{c|}{MUXORNet-11} \\
    \cline{3-6}
     &   & FP & Bin. &   FP  & Bin. \\
\hline
Baseline gamma-inversed &  \textit{32-b}  &   82.29  & 79.44     &  83.03   &  78.44 \\ \hline
\multirow{3}{*}{\makecell{Fixed Linear \\ Thermometer \cite{FracBNN}}} & 8 & 76.51    &  73.40    &   78.39    &  74.99 \\ 
  &  16  & 77.99   & 75.09 & 80.81  & 76.98 \\
  &  32  & 78.86   & 75.86 & 80.85  & 77.30 \\
\hline
 \multirow{3}{*}{\makecell{\textbf{Ours} \\ GLT}} & 8 & 78.24   & 75.69  & 80.90  & 77.45 \\
  &  16 &  79.54   & 76.46   &  81.78  &  78.05  \\
   & 32 &  80.35   & 77.88   &  82.06  &  \textbf{78.51}  \\
\hline
\end{tabular}
\label{encoding_benchmark_gamma_inversed}
\vspace{-4mm}
\end{table}


\subsection{Validation of the block pruning on fully-binarized model}
\begin{figure}[h]
     \centering
         \begin{subfigure}[b]{0.24\textwidth}
        \includegraphics[scale = 0.305]{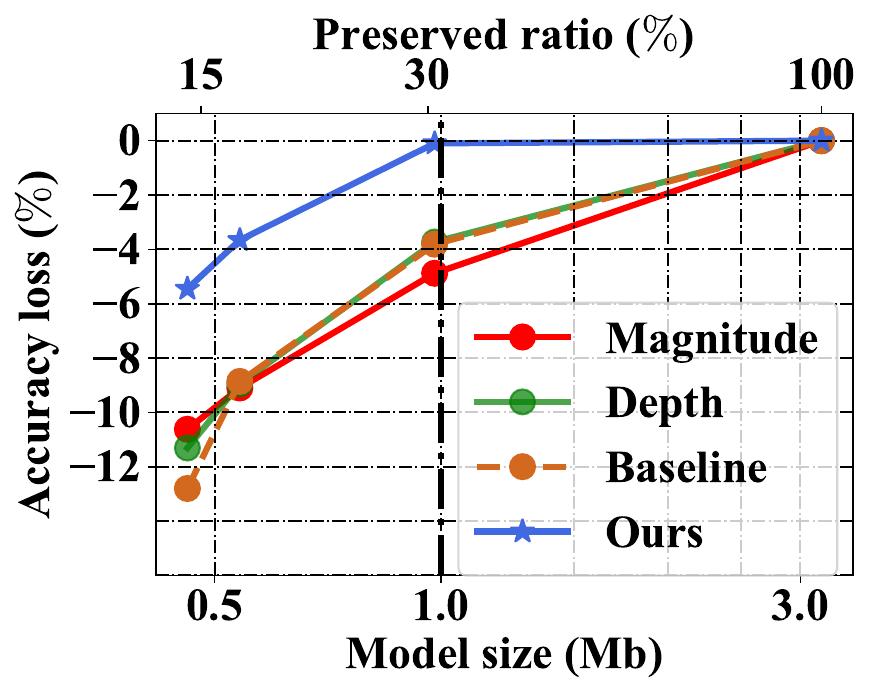} 
        \label{model_size_acc}
    \end{subfigure}
    \hspace{-2mm}
    \begin{subfigure}[b]{0.24\textwidth}
        \includegraphics[scale = 0.305]{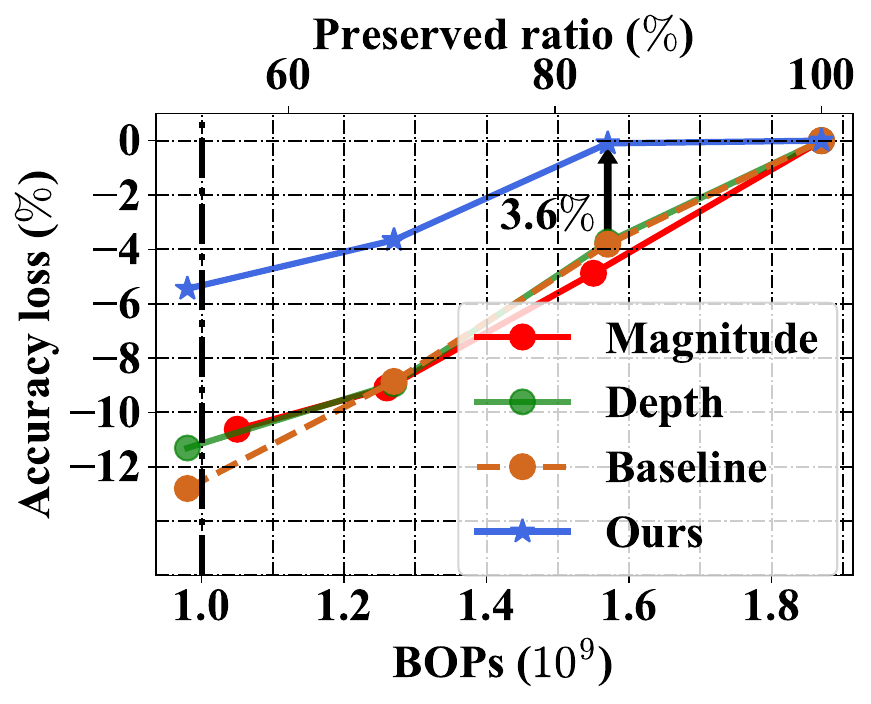} 
        \label{bops_acc}
    \end{subfigure}
    \vspace{-6mm}
     \caption{Trade-off curve for model size/BOPs reduction and accuracy loss of pruned MUXORNet-11 (original acc: 78.5$\%$).}
     \label{pruning_curves}
    \vspace{-5mm}
\end{figure}

In this part, we consider the MUXORNet-11 model with GLT and $M=32$, trained on gamma-inversed dataset ($78.51\%$ acc.), to conduct experiment. The pre-trained model after the first stage in~\ref{val_glt} is used as teacher in our KD scheme. We set the groups $g$ as $1, 2, 8$ for three blocks, respectively, since the last layers are less sensitive to model's performance. We compare our gradual block pruning method with three competitors, involving 1) \textbf{baseline}: the pruned model but trained from scratch; 2) \textbf{depth}: depth pruning \cite{Depthpruning} using our LWCs as auxiliary model but trained in one shot without the KD loss $\mathcal{L}_{distr}$; 3) \textbf{magnitude}: channel-pruning based on the magnitude of the latent weights (inspired by \cite{10044856}) in which the pruning ratio is computed to have the same model size as the block-pruned model. The pruned models are trained during $300$ epochs with learning rate initialized at $10^{-3}$ and reduced to $10^{-10}$ using cosine decay scheduler. Figure~\ref{pruning_curves} shows the trade-off of model size/BOPs-accuracy loss of the pruned models. At each pruning point, our method always offers more than $3.6\%$ higher accuracy compared to other methods. In particular, when pruning the third block, we successfully reduce the model size by $70\%$ and the number of BOPs \cite{BOP} by $16\%$, this nearly without accuracy drop. Even if we seek for a model of under $0.5$Mb and $1$GBOPs, our method still reaches $73\%$ accuracy while other methods cannot exceed $68\%$. These results demonstrate the effectiveness of our method on designing extremely tiny, fully-binarized models.

\vspace{-0.1cm} 

\section{Conclusion}
This work addresses the end-to-end design of small-sized, fully-binarized networks for always-on inference use cases. To do this, we first introduce the GLT encoding scheme to properly manage input images binarization, jointly with performing a trained global tone mapping. Then, we propose a gradual block-based pruning strategy combined with a distributional loss to limit the overall accuracy drop. Experimental results on VWW and STL-10 demonstrate that GLT enables an efficient training stage, with a better generalization on unseen dataset. Our methods enable highly accurate models ($\sim$78.5\% accuracy on STL-10) that require 1-bit computations only with a model size under $1$Mb, while relying on inputs of 32 bit-planes. Future works may investigate the performance on practical in-sensor data (\eg, on mosaiced frames impaired by fixed pattern noise and dead pixels), enabling ISP-free but still highly accurate always-on inference modules. 

\newpage 

\bibliographystyle{IEEEtran}
\input{AICAS25.bbl}

\end{document}